\documentclass[letterpaper, 10 pt, conference]{ieeeconf}  

\usepackage{amsmath,amssymb}
\usepackage{graphicx}
\usepackage{booktabs}
\usepackage{multirow}
\usepackage{makecell}
\usepackage[table]{xcolor}
\usepackage{array}
\usepackage{cite}
\usepackage{algpseudocode}
\newcounter{algorithmblock}

\definecolor{condblue}{RGB}{218,227,243}
\definecolor{oursblue}{RGB}{240,244,250}
\definecolor{ppopink}{RGB}{252,228,236}

\newcommand{\hc}[2]{\cellcolor{#1}\hspace{1.5pt}#2\hspace{1.5pt}}

\makeatletter
\let\NAT@parse\undefined
\makeatother

\usepackage[hidelinks]{hyperref}

\IEEEoverridecommandlockouts                              

\title{\LARGE \bf
Tactile-Based Jamming-Preventive Precision Peg-in-Hole Assembly
}

\author{Yijia Lei, Qiyin Huang, Tiemin Li, and Yao Jiang%
\thanks{All authors are with Tsinghua University, Beijing, China.}%
\thanks{Corresponding author: Yao Jiang.}%
}

\begin{document}

\maketitle
\thispagestyle{empty}
\pagestyle{empty}


\begin{abstract}
Precision peg-in-hole assembly under sub-millimeter clearances is highly susceptible to jamming, where commanded hand motion progressively loses effective transmission into actual peg insertion. We formulate jamming prevention as online regulation of motion transmission. From bilateral in-hand tactile fields, we construct a simple low-dimensional force--position coupled tactile cue whose components capture directional contact deviation and the associated moment tendency. These cues directly serve as tactile control coordinates: their signs determine lateral correction directions, while their action-induced evolution regulates correction authority during maintained axial advancement. Real-robot experiments across multiple peg shapes, sizes, and sub-millimeter clearances show consistently higher success than stage-specific PPO baselines. Tactile-cue validation and ablation further demonstrate physically consistent cue behavior, improved insertion stability, and reduced motion-transfer loss. These results show that simple tactile control coordinates and their action--response evolution provide an effective basis for jamming prevention in precision insertion.
\end{abstract}


\section{INTRODUCTION}

Precision peg-in-hole assembly is a canonical yet challenging contact-rich manipulation task in autonomous assembly. Tight clearances make insertion highly sensitive to residual peg--hole mismatch. Even a small mismatch can produce asymmetric contact at the hole rim or wall. As the robot continues to advance, lateral contact, friction, fingertip deformation, or relative motion within the grasp can progressively reduce the conversion of commanded hand motion into actual peg insertion. In classical peg--hole mechanics, jamming is characterized as a contact equilibrium in which the applied insertion wrench is balanced by contact reactions, preventing further insertion~\cite{whitney1982quasistatic,li2025advances}. The progressive motion-transfer loss described above is therefore regarded as an operational precursor to such a jammed state. Preventing jamming therefore requires more than detecting a final blocked state: the controller must determine, during contact evolution, whether nominal advancement is still producing true insertion.

The central control challenge is to maintain effective axial advancement while relieving residual peg--hole mismatch. Corrective-motion effectiveness depends on the current contact constraints and compliance~\cite{whitney1982quasistatic,chen2025robust}, while peg--environment forces can alter the in-hand pose under compliant grasping~\cite{gibbons2023tactile}. An ineffective correction may therefore be accommodated by constrained contact or in-hand motion rather than relieving the mismatch. It must consequently be determined whether the resulting contact change supports further correction.

\begin{figure}[t]
    \centering
    \includegraphics[width=\columnwidth]{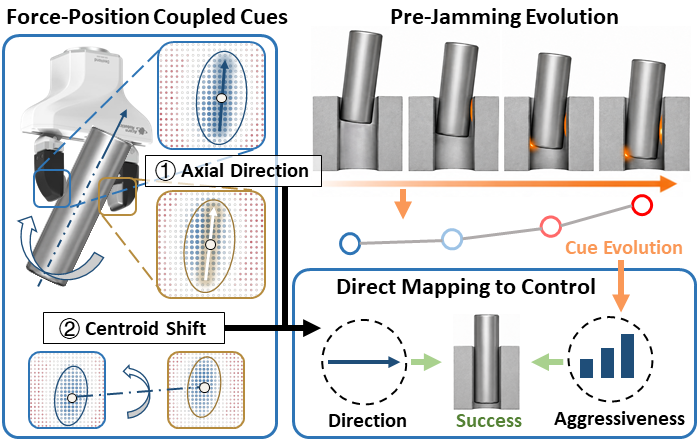}
    \caption{\textbf{Key concept of tactile-based jamming-preventive precision insertion.}Bilateral tactile fields are reduced to force--position coupled cues that represent pre-jamming contact evolution. Cue signs determine correction direction, while action-induced cue evolution regulates correction authority during maintained axial advancement.}
    \label{fig:teaser}
    \vspace{-5mm}
\end{figure}

Insertion failures have been addressed through model-based contact reasoning, force or compliance control, and learned policies~\cite{xu2019compare,li2025advances}. Geometry- or state-based methods rely on predefined models or contact modes, force reduction does not necessarily indicate improved alignment, and learned policies may degrade outside their training distributions. An important question therefore remains: how can action-induced contact evolution be used to evaluate correction effectiveness and regulate continued advancement online?

To address this question, tactile feedback is required to characterize both the current contact deviation and the outcome of the recent action. Based on this principle, precision insertion is formulated as tactile regulation of motion transmission before jamming. A low-dimensional force--position coupled tactile cue is constructed from bilateral in-hand tactile fields. Its principal-direction component describes directional contact deviation. Its centroid-shift component reflects the displacement of the contact line of action and the associated moment tendency. These components are used to determine the correction direction, and their evolution is monitored after each correction. A retained decrease indicates that the correction has improved the contact condition. Stagnation or rebound indicates that further correction may be ineffective or destabilizing. This action-induced evolution is used to regulate lateral correction while axial advancement is maintained. A closed loop between tactile sensing, correction evaluation, and motion regulation is thereby established. Full peg--hole pose reconstruction, contact-state enumeration, and task-specific offline training are not required.

The proposed method is evaluated in real-robot peg-in-hole tasks with multiple shapes, sizes, and sub-millimeter clearances. 
The proposed method is compared with stage-specific PPO policies that likewise perform closed-loop control from low-dimensional states derived from bilateral tactile feedback. A cue-validation study examines the consistency between the tactile cues and offline visual and force-related references. An ablation study compares the proposed dynamic mechanism with fixed-regulation variants. The results show improved task success, reduced motion-transfer loss, and more effective conversion of commanded axial motion into actual peg insertion.

The main contributions of this work are summarized as follows:
\begin{enumerate}
    \item We formulate jamming prevention as online regulation of motion transmission. Based on this formulation, we propose a force--position coupled tactile cue that captures directional contact deviation and the associated moment tendency.

    \item We propose a tactile-cue-guided correction-regulation mechanism during axial advancement. It evaluates correction effectiveness from action-induced tactile evolution and regulates correction during axial advancement.

    \item We validate the proposed tactile cue and control mechanism in real-robot tasks across multiple peg shapes, sizes, and sub-millimeter clearances. The results demonstrate improved insertion success and reduced motion-transfer loss without object-specific geometric modeling or task-specific offline training.
\end{enumerate}

\section{Related Work}

\subsection{Mechanics of Jamming and Motion-Transfer Failure}

Classical jamming analyses determine insertion feasibility from contact configuration, friction, reaction forces, and force/moment equilibrium~\cite{fei2004contact,sathirakul1998jamming,wu2022robot}. Two-point contact and unfavorable compliance can produce resistance or wedging, while appropriate compliance can mitigate these failures~\cite{zhang2019peg,goli2024jamming}. These mechanics motivate treating motion-transfer loss as an operational manifestation of increasingly unfavorable constrained contact before a jammed state is reached.

Building on this mechanical understanding, we interpret motion-transfer failure as an operational manifestation of increasingly unfavorable constrained contact: commanded motion becomes progressively less effectively transmitted into the intended peg motion. This view does not introduce a new contact state, but connects classical jamming mechanics to the control problem of evaluating whether corrective actions are improving the contact condition during insertion.

\subsection{Jamming Mitigation and Insertion Control}

Existing insertion controllers mitigate jamming through contact-state or geometry-based reasoning, force or compliance regulation, and learned policies. Contact-state methods infer assembly modes from force or torque observations and switch control strategies accordingly~\cite{lau2003hidden,jasim2017contact}, while geometry- and configuration-based approaches exploit motion constraints, contact mechanics, or predefined contact manifolds to guide search and insertion~\cite{li2021robotic,chen2025robust}. Force and compliance methods regulate contact loads or impedance to accommodate constrained interaction~\cite{zhang2019peg,goli2024jamming}, and learning-based approaches learn insertion actions, guiding forces, or impedance behaviors from interaction data~\cite{tang2025visual,zang2025safe,yang2021learning}. These approaches provide complementary mechanisms for selecting or regulating corrective actions. However, comparatively less attention has been paid to explicitly using the contact evolution produced by the latest correction to determine whether that correction has produced and retained a favorable change, and whether subsequent correction should therefore be maintained or attenuated during continued advancement.

\subsection{Tactile Representation for Contact Evolution and Control}

High-resolution tactile sensing has enabled structured representations of contact for manipulation and assembly. Prior work has used tactile observations for in-hand pose estimation and localization~\cite{li2014localization}, tactile-feedback insertion and pose compensation~\cite{gibbons2023tactile}, tactile-based active inference for force-controlled insertion~\cite{kamijo2026tactile}, extrinsic contact localization~\cite{kim2022active}, interactive part-mating perception~\cite{ota2023tactilefilter}, and closed-loop pose feedback~\cite{ota2023tactile}. Vision-based tactile sensors such as Tac3D further provide distributed deformation and force information, allowing a single tactile field to encode contact geometry, spatial deformation, and load-related responses~\cite{zhang2022tac3d}. These studies establish tactile sensing as an effective source of state information and feedback for contact-rich control. Our focus is complementary: rather than using tactile feedback only to estimate the current pose or contact state, we summarize the bilateral tactile fields with low-dimensional directional and centroid-shift cues and use their action-induced temporal evolution to evaluate the outcome of each correction. Tactile feedback therefore serves not only to characterize the current contact, but also to determine whether a corrective action has produced and retained a favorable contact change.

\section{Method}

\begin{figure}[t]
    \vspace{3mm}
    \centering
    \includegraphics[width=\columnwidth]{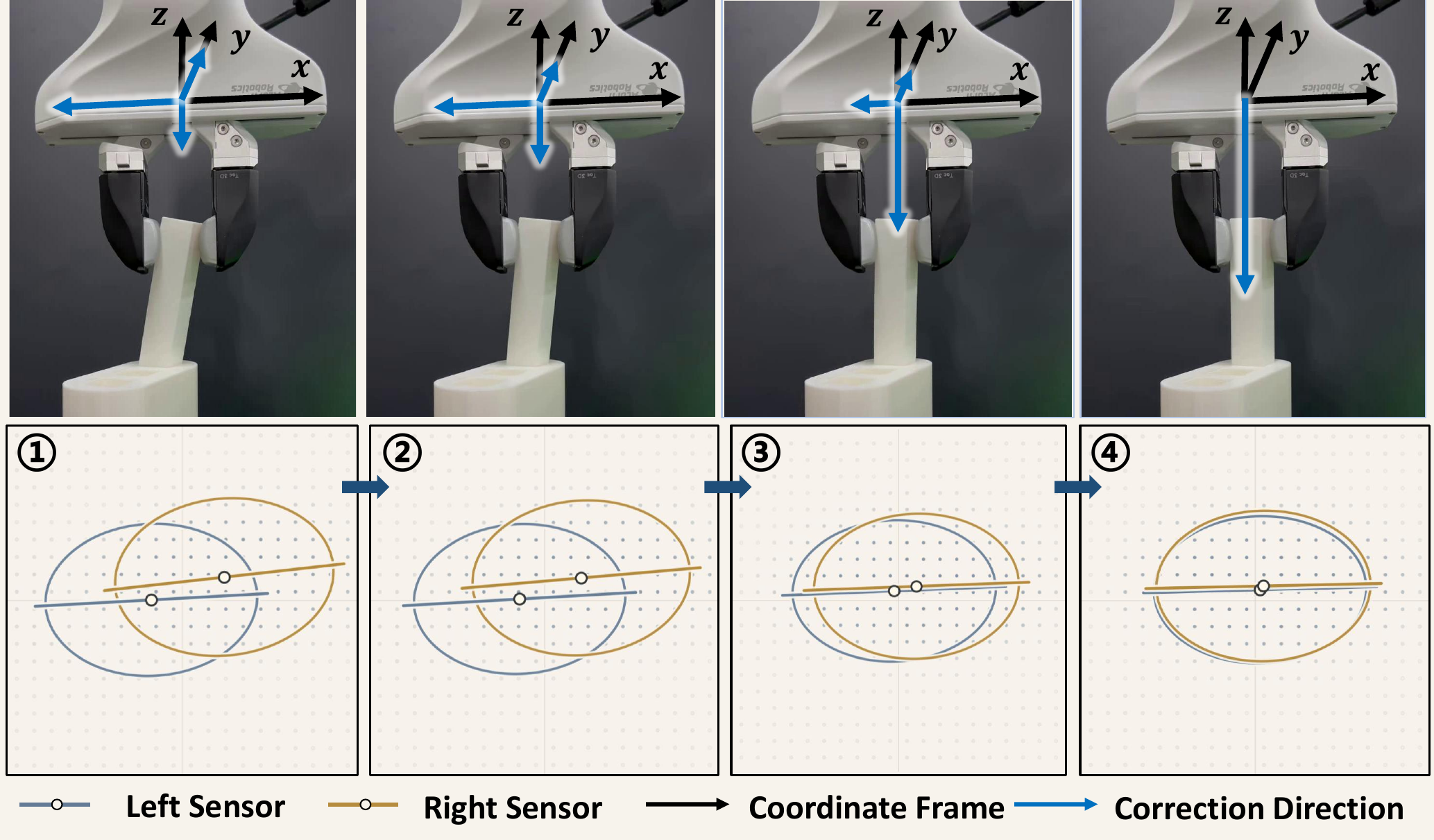}
    \caption{\textbf{Correspondence between tactile-field evolution and corrective control.}
    (a) Representative insertion states with task-space coordinate axes (gray) and applied correction directions (blue).
    (b) Bilateral tactile fields and the corresponding centroid-shift cue \(e_x\) and directional cue \(e_\theta\) over \(t_1\)--\(t_4\).}
    \label{fig:tactile_cue}
    \vspace{-5mm}
\end{figure}

\subsection{Task Setting}

We consider unchamfered, flat-ended peg-in-hole tasks under sub-millimeter clearances. Clearance is defined as the hole--peg size difference: diameter difference for cylindrical pegs and minimum across-flats difference for prismatic pegs. A Franka Panda equipped with a two-finger gripper and two fingertip Tac3D sensors is used~\cite{zhang2022tac3d}. The peg is grasped between compliant fingertips; the hole is fixed, and only the nominal insertion direction is assumed known.

At each control step \(t\), an incremental control command is executed in task space,
\begin{equation}
\mathbf{u}_t =
[\Delta x_t,\Delta y_t,\Delta z_t,
\Delta \mathrm{roll}_t,\Delta \mathrm{pitch}_t,\Delta \mathrm{yaw}_t]^\top ,
\end{equation}
where \(\Delta z_t\) denotes axial advancement and \(\Delta x_t,\Delta y_t\) denote lateral correction. The rotational commands are stage dependent. During hole-rim alignment, only yaw is actively corrected, while roll and pitch commands are set to zero. During in-hole insertion, all three rotational commands are set to zero. Small uncommanded angular mismatch can nevertheless be passively accommodated by the compliant tactile fingertips instead of being rigidly imposed by the robot. The gripper orientation remains fixed with respect to the world frame during in-hole insertion, providing a consistent task-aligned reference direction for the tactile cues defined below.

The coordinate convention consists of the world frame \(\mathcal{F}_W\), the hand frame \(\mathcal{F}_H\), and two sensor-local frames \(\mathcal{F}_S^L,\mathcal{F}_S^R\). For each sensor \(s\in\{L,R\}\), a \(20\times20\) three-dimensional deformation field and force field are provided by Tac3D at time step \(t\). Let
\(\mathbf{p}_i=[u_i,v_i]^\top\) denote the two-dimensional grid coordinate of the \(i\)-th taxel, and let its deformation and force vectors be
\begin{equation}
\mathbf d^{(s)}_{t,i}
=
[\delta x^{(s)}_{t,i},
 \delta y^{(s)}_{t,i},
 \delta z^{(s)}_{t,i}]^\top ,
\qquad
\mathbf f^{(s)}_{t,i}
=
[f^{(s)}_{x,t,i},
 f^{(s)}_{y,t,i},
 f^{(s)}_{z,t,i}]^\top .
\end{equation}
The complete deformation and force fields are denoted by
\(\mathbf D_t^{(s)}\) and \(\mathbf F_t^{(s)}\), respectively. Tac3D obtains its calibrated force field from the measured deformation through a sensor-specific stiffness mapping; hence the deformation and force interfaces provide complementary representations of the same contact response.

Rather than being used to reconstruct the full peg--hole relative pose, these tactile fields are regarded as online observations of the in-hand contact configuration. Since the peg is grasped by compliant fingertips, environmental contact, friction, and local collision can cause small in-hand motion of the peg and continuous changes in the bilateral tactile fields. Therefore, the tactile fields contain contact position, local geometry, and load-related deformation responses.

Residual peg--hole mismatch is used to denote contact-relevant mismatch that disrupts the conversion from nominal advancement to actual insertion. Its tactile manifestations include directional geometric deviation and the associated shift in the contact line of action and moment tendency. Instead of reconstructing the full six-dimensional peg--hole pose, two low-dimensional tactile cues are extracted, and their online evolution is used to evaluate the contact response to correction.

\subsection{Force--Position Coupled Tactile Cue}

During tight-clearance insertion, residual peg--hole mismatch changes both the dominant direction of the in-hand deformation and the locations at which the bilateral contact responses act. The resulting representation is referred to as force--position coupled because the calibrated deformation field contains a load-related response while preserving its spatial distribution; no exact force or moment is directly estimated.

For each sensor \(s\in\{L,R\}\), the magnitude of the normal deformation is used as the local contact-response weight,
\[
w^{(s)}_{t,i}
=
\left|\delta z^{(s)}_{t,i}\right|,
\]
from which the weighted contact centroid and covariance are computed:
\[
\begin{aligned}
\boldsymbol{\mu}^{(s)}_t
&=
\frac{\sum_i w^{(s)}_{t,i}\mathbf p_i}
     {\sum_i w^{(s)}_{t,i}},
\\
\boldsymbol{\Sigma}^{(s)}_t
&=
\frac{
\sum_i w^{(s)}_{t,i}
(\mathbf p_i-\boldsymbol{\mu}^{(s)}_t)
(\mathbf p_i-\boldsymbol{\mu}^{(s)}_t)^\top}
{\sum_i w^{(s)}_{t,i}}.
\end{aligned}
\]
Here, \(\boldsymbol{\mu}^{(s)}_t\) describes the acting position of the contact response, while \(\boldsymbol{\Sigma}^{(s)}_t\) describes its dominant spatial direction. The eigenvector corresponding to the largest eigenvalue is denoted by \(\mathbf r^{(s)}_{1,t}\).

Let \(\theta_t^{(s)}\) denote the signed axial angle of this principal direction with respect to the task-aligned tactile reference direction. Because the hand orientation is fixed during in-hole insertion, this reference direction remains fixed with respect to the nominal insertion direction. After transforming the right principal direction into the common bilateral coordinate convention, its \(\pi\)-periodic ambiguity is resolved by selecting
\[
\widetilde{\theta}^{(R)}_t
=
\theta^{(R)}_t+k_t\pi,
\qquad
k_t
=
\operatorname*{arg\,min}_{k\in\mathbb Z}
\left|
\theta^{(R)}_t+k\pi-\theta^{(L)}_t
\right|.
\]
The bilateral directional cue is then
\[
e_{\theta,t}
=
\frac{1}{2}
\left(
\theta^{(L)}_t+
\widetilde{\theta}^{(R)}_t
\right).
\]
Its sign determines the associated correction direction, while
\(\lvert e_{\theta,t}\rvert\) measures the magnitude of the directional deviation.

The second component describes the relative shift of the bilateral contact centroids along the moment-sensitive direction. Let
\(\mu^{(s)}_{\perp,t}\) denote this centroid component. The centroid-shift cue is defined as
\[
e_{x,t}
=
\mu^{(R)}_{\perp,t}
-
\mu^{(L)}_{\perp,t},
\]
which corresponds to \(c_{Rx}-c_{Lx}\) in the controller.

Its moment sensitivity can be understood from a reduced bilateral-contact model. The contact moment is generated by
\[
\boldsymbol{\tau}
=
\mathbf r_L\times\mathbf F_L
+
\mathbf r_R\times\mathbf F_R.
\]
Under an approximately stable and bilaterally balanced grasp preload,
\[
\|\mathbf F_L\|
\approx
\|\mathbf F_R\|
\approx F,
\]
so the relevant moment component varies, to first order, as
\[
\tau_{\mathrm{rel}}
\propto
F
\left(
\mu^{(R)}_{\perp,t}
-
\mu^{(L)}_{\perp,t}
\right)
=
F e_{x,t}.
\]
Therefore, when the bilateral preload varies slowly over the correction interval, \(e_{x,t}\) is expected to vary monotonically with the corresponding moment tendency. It is used as a moment-sensitive proxy rather than a numerical moment estimate.

Finally, the in-hole tactile cue is
\[
\mathbf e_t
=
[e_{x,t},e_{\theta,t}]^\top .
\]
Correction directions are determined by the signed cue values, whereas correction effectiveness is evaluated from their magnitudes. A retained decrease in \(\lvert e_{j,t}\rvert\) indicates that the current correction improves the contact condition; stagnation or rebound indicates that further correction may become ineffective or destabilizing.

\subsection{Tactile-Cue-Guided Correction Regulation During Axial Advancement}

The two tactile components are treated as low-dimensional tactile control coordinates. Each component \(j\in\{x,\theta\}\) is associated with a fixed lateral correction basis \(\mathbf b_j\) in task space, determined by the sensor--hand mounting geometry. The correction direction is selected according to its signed value through a dead-zone rule,
\begin{equation}
s_{j,t}
=
-\operatorname{sgn}_{\epsilon_j}(e_{j,t}),
\end{equation}
where \(\epsilon_j\) suppresses small-amplitude tactile noise. Bounded lateral micro-steps are then applied by the low-level executor in the form
\[
\Delta\mathbf p^{\mathrm{lat}}_{j,t}
=
a_{j,t}s_{j,t}\mathbf b_j,
\qquad
0\leq a_{j,t}\leq a_j^{\max}.
\]
During in-hole insertion, no rotational command is actively applied, while axial advancement remains continuously active. The exact bilateral coordinate convention and cue-to-Cartesian basis mappings are fixed by the hardware configuration and are provided in the supplementary material.

Importantly, the cue magnitude is not converted into a metric Cartesian displacement. Instead, the cue magnitude is used to define a tactile-space servo target. For each component, a target residual ratio \(\rho_{j,t}\) is maintained, and within one macro step the cue magnitude is driven from
\(\lvert e^{\mathrm{pre}}_{j,t}\rvert\)
toward
\[
\lvert e^{\mathrm{tar}}_{j,t}\rvert
=
\rho_{j,t}
\lvert e^{\mathrm{pre}}_{j,t}\rvert
\]
through bounded micro-corrections along the selected channel. Only a locally consistent cue--action relation is therefore required: a favorable correction along \(\mathbf b_j\) should reduce the corresponding cue magnitude. A smaller \(\rho_{j,t}\) corresponds to more aggressive lateral correction, while a larger value produces a more conservative correction.

During each macro step, the best intermediate cue magnitude
\(\lvert e^{\mathrm{best}}_{j,t}\rvert\)
and the terminal magnitude
\(\lvert e^{\mathrm{post}}_{j,t}\rvert\) are recorded, with the latter measured after a short settling observation. Their completion scores are
\begin{equation}
\eta^{\mathrm{best}}_{j,t}
=
\operatorname{clip}_{[0,1]}
\left(
\frac{
\lvert e^{\mathrm{pre}}_{j,t}\rvert
-
\lvert e^{\mathrm{best}}_{j,t}\rvert
}{
(1-\rho_{j,t})
\lvert e^{\mathrm{pre}}_{j,t}\rvert+\varepsilon
}
\right),
\end{equation}
and
\begin{equation}
\eta^{\mathrm{post}}_{j,t}
=
\operatorname{clip}_{[0,1]}
\left(
\frac{
\lvert e^{\mathrm{pre}}_{j,t}\rvert
-
\lvert e^{\mathrm{post}}_{j,t}\rvert
}{
(1-\rho_{j,t})
\lvert e^{\mathrm{pre}}_{j,t}\rvert+\varepsilon
}
\right).
\end{equation}
The final completion score is
\begin{equation}
\eta_{j,t}
=
\min
\left(
\eta^{\mathrm{best}}_{j,t},
\eta^{\mathrm{post}}_{j,t}
\right).
\end{equation}
Under this min-rule, an improved state must both be reached during execution and be retained in the terminal observation. If the terminal cue rebounds beyond the best intermediate value by more than a prescribed margin \(\gamma\), \(\eta_{j,t}\) is reset to zero.

To reduce the effect of single-step noise, the completion score is filtered by
\begin{equation}
\bar{\eta}_{j,t}
=
\alpha\bar{\eta}_{j,t-1}
+
(1-\alpha)\eta_{j,t},
\end{equation}
where \(\alpha\in(0,1)\). The target residual ratio is selected from a discrete schedule,
\begin{equation}
\rho_{j,t}
=
\rho_{\mathrm{grid}}(I_{j,t}),
\qquad
I_{j,t+1}
=
\min
\left\{
I_{j,t}
+
\Delta I(\bar{\eta}_{j,t}),
I_{\max}
\right\},
\end{equation}
where lower completion scores produce larger nonnegative index increments. A conservative update is enforced when rebound is detected. Thus, when a tactile component can still be effectively reduced, the current correction authority is retained; when improvement becomes insufficient or unstable, \(\rho_{j,t}\) is increased and subsequent lateral correction is reduced.

The monotonic schedule is a deliberate consequence of the contact geometry during tight-clearance insertion. As peg--hole engagement increases, the surrounding hole constraint generally becomes more restrictive and the admissible range for lateral correction contracts. Correction authority is therefore allowed to be preserved or progressively reduced, but not restored to an earlier aggressive level after a more conservative regime has been reached. Responsiveness to newly observed mismatch is not prevented by this one-way schedule: the current signed tactile cues continue to update the correction directions, while the correction magnitude remains bounded by the most conservative authority reached so far. In this way, the direction of scheduling is determined by the task geometry, whereas the timing and rate of transition toward more conservative correction are determined by the measured action-induced tactile evolution.

The axial command itself is not dynamically scheduled: axial advancement remains continuously active, while lateral correction authority is regulated by the tactile scheduler. Increasing \(\rho_{j,t}\) therefore attenuates lateral correction while leaving axial advancement unchanged. For the fixed-regulation variants, \(\rho_{j,t}\) is held constant throughout insertion.

\subsection{Control Instantiation for Precision Insertion}

The controller is instantiated in two consecutive regimes: hole-rim contact and in-hole insertion. Although the two regimes use stage-specific tactile observables under different contact constraints, both regulate correction from the tactile evolution induced by the correction itself.

\begin{figure}[t]
\vspace{3mm}
\centering
\begin{minipage}{0.98\columnwidth}
\refstepcounter{algorithmblock}
\label{alg:overall}

\small
\noindent\textbf{Algorithm 1:}\\
\textbf{Tactile-Cue-Guided Jamming-Preventive Insertion}

\vspace{0.5mm}
\hrule
\vspace{0.8mm}

\begin{algorithmic}[1]
\Require Bilateral tactile fields and nominal insertion direction

\Statex \textbf{Rim-contact regulation}
\State Apply translational probe and evaluate \(q_t\)
\State Infer response polarity \(d\) from three valid samples of \(q_t\)
\Repeat
    \State Apply yaw correction according to \(d\) and update \(q_t\)
\Until{\(d(q_t-q_{t^-})<-\delta_q\)}

\Statex \textbf{In-hole correction regulation}
\While{insertion is active}
    \State Extract tactile control coordinates
    \Statex \hspace{\algorithmicindent}
    \(\mathbf e_t=[e_{x,t},e_{\theta,t}]^\top\)
    \State Set correction directions
    \Statex \hspace{\algorithmicindent}
    \(s_{j,t}\gets-\operatorname{sgn}_{\epsilon_j}(e_{j,t})\),
    \(j\in\{x,\theta\}\)
    \State Set tactile residual targets
    \Statex \hspace{\algorithmicindent}
    \(\lvert e^{\mathrm{tar}}_{j,t}\rvert
    \gets\rho_{j,t}\lvert e^{\mathrm{pre}}_{j,t}\rvert\)
    \State Maintain axial advancement with bounded lateral correction
    \State Observe best and post-settling cue magnitudes
    \State Evaluate retained correction effectiveness \(\eta_{j,t}\)
    \State Apply the rebound gate and update \(\bar{\eta}_{j,t}\)
    \State Update \(\rho_{j,t}\) monotonically from correction effectiveness
\EndWhile
\end{algorithmic}

\vspace{0.5mm}
\hrule
\vspace{-5mm}
\end{minipage}
\end{figure}

\paragraph{Rim-stage tactile probing and entry regulation.}
At the hole rim, a bilateral moment-related response cue is constructed from the distributed Tac3D taxel positions and forces,
\[
q_t
=
\sum_{s\in\{L,R\}}
\sum_i
\left[
\widetilde{\mathbf r}_{t,i}^{(s)}
\times
\widetilde{\mathbf f}_{t,i}^{(s)}
\right]_{\mathrm{rel}},
\]
where \(\widetilde{\mathbf r}_{t,i}^{(s)}\) denotes the effective moment arm under the bilateral sensor convention and current gripper opening, and \(\widetilde{\mathbf f}_{t,i}^{(s)}\) is the corresponding taxel force. The resulting \(q_t\) is treated as a moment-related response cue rather than a numerical torque estimate.

A short translational probing motion is first applied, and the latest three valid samples of \(q_t\) are used to establish a local response polarity \(d\in\{-1,+1\}\): \(d=+1\) when both adjacent cue increments exceed a small positive threshold, and \(d=-1\) when both are negative beyond the corresponding threshold. The subsequent yaw-correction direction is determined by this polarity. During yaw correction, a sufficiently large response reversal,
\[
d\left(q_t-q_{t^-}\right)<-\delta_q,
\]
indicates that the current correction is no longer locally favorable; the correction is then stopped and the controller switches to the in-hole stage.

\paragraph{In-hole correction regulation.}
After the transition, all active rotational corrections are removed, and the regulation in Sec.~III-C is applied to \(e_{x,t}\) and \(e_{\theta,t}\) while maintaining axial advancement.

Thus, response reversal terminates rim-stage correction, while retained cue reduction and rebound regulate in-hole correction under the same action--response principle.

\section{EXPERIMENT}

\subsection{Experimental Setup and Evaluation Protocol}

The proposed jamming-suppression strategy is validated on a real robotic platform. The experimental platform consists of a Franka Emika Panda arm, a two-finger parallel gripper, and Tac3D tactile sensors mounted on both fingertips. In all closed-loop insertion experiments, Tac3D is the only online observation modality; external vision, the true peg--hole relative pose, and manually annotated contact states are not used by the controller. The external industrial camera and ArUco markers are used only for offline analysis: an ArUco-based directional trend reference is provided in the tactile-cue validation experiment, while the axial displacement of the peg relative to the hand is measured in the ablation experiment. Neither visual signal is used for closed-loop control.

The method is evaluated with three peg shapes, namely cylindrical, square-prismatic, and triangular-prismatic pegs, two sizes, \(20\,\mathrm{mm}\) and \(16\,\mathrm{mm}\), and three sub-millimeter clearances, \(0.5\,\mathrm{mm}\), \(0.4\,\mathrm{mm}\), and \(0.3\,\mathrm{mm}\), forming \(3\times2\times3=18\) real-robot conditions. Three independent protocols are evaluated for each condition. Rim-entry and end-to-end trials are initialized from a challenging rim-contact configuration generated by an upstream coarse-alignment procedure: the lowest point of the flat peg end face lies below the hole-rim plane, while the end face remains partially, and in some trials substantially, outside the aperture, with the peg body tilted by approximately \(20^\circ\)--\(30^\circ\). For the in-hole protocol, the peg tip is first inserted slightly into and kept fully inside the hole; the peg tail is then horizontally displaced by a randomized offset of at most \(15\,\mathrm{mm}\), subject to the geometric constraint that the tip remains inside the aperture. Initializations violating this constraint are discarded and resampled before the trial begins.

The rim-entry, in-hole, and end-to-end evaluations use independent trial sets rather than different labels of the same executions. For each method, condition, and protocol, \(20\) independent real-robot trials are conducted. For PPO end-to-end evaluation, runs with premature or belated switching by the auxiliary transition detector were treated as invalid protocol execution and repeated; such cases accounted for less than 5\% of attempts and were excluded before success or failure counting. The same controller parameters are used for all shapes, sizes, and clearances. Both the rim and in-hole stages are capped at \(200\) control steps; in the end-to-end protocol, a separate \(200\)-step budget is assigned to each stage. Reaching the corresponding step limit terminates the stage and is counted as task failure. Trials involving object release are treated as immediate failures and are manually terminated for safety. For the in-hole and end-to-end protocols, insertion is considered complete when either the hand reaches a pre-calibrated terminal height corresponding to full insertion in the absence of motion-transfer loss, or the axial resistance reaches a preset threshold equal to the nominal grasp-force level. Full seating is verified after gripper release by confirming that the peg cannot be advanced further.

\begin{figure}[t]
    \vspace{3mm}
    \centering
    \includegraphics[width=\columnwidth]{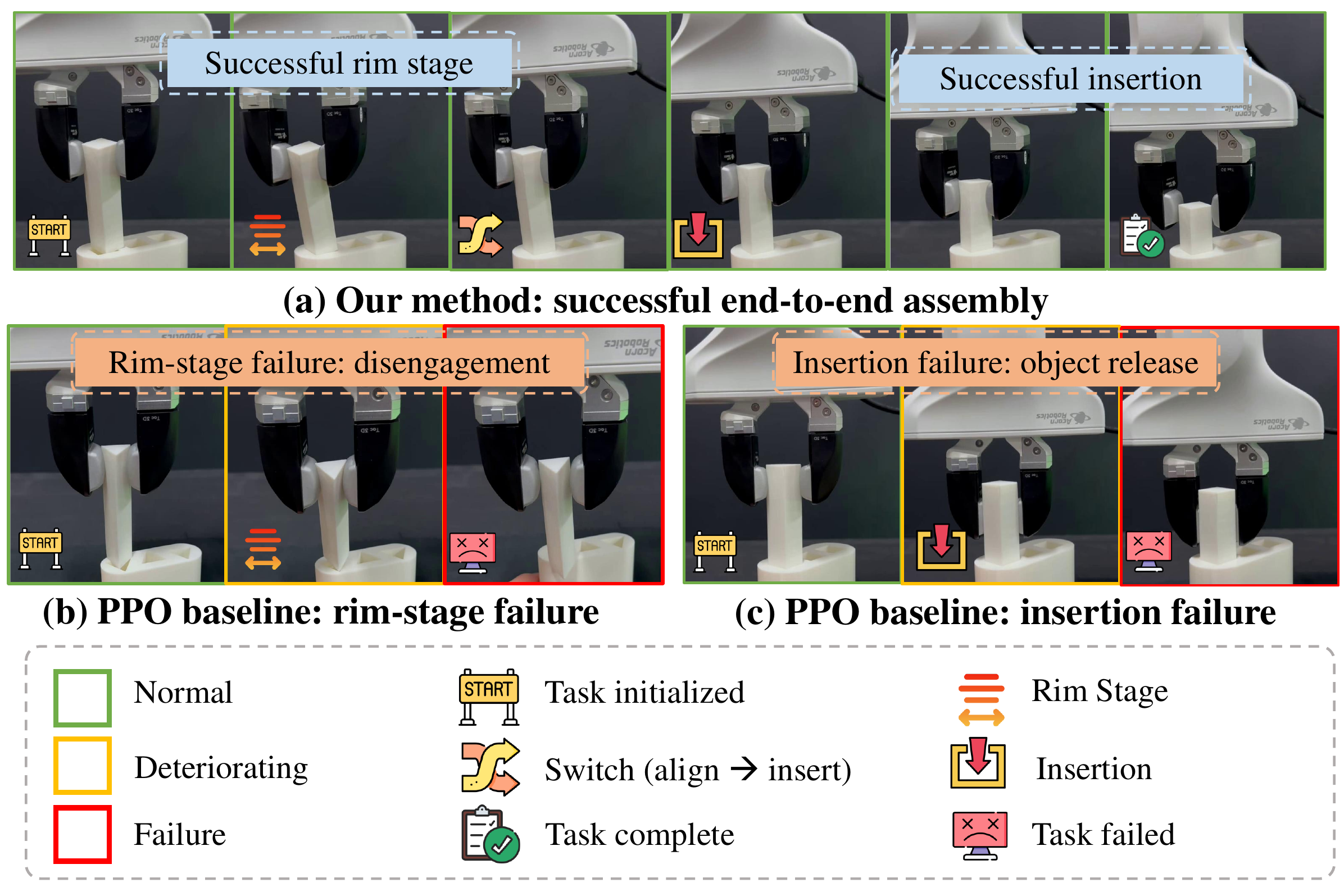}
    \caption{\textbf{Representative end-to-end executions.}
    (a) Successful assembly with the proposed method.
    (b) PPO rim-stage disengagement.
    (c) PPO insertion-stage object release.}
    \label{fig:exp_results}
    \vspace{-5mm}
\end{figure}

PPO is adopted as a learning-based tactile-control baseline. To provide sufficient control capability under the distinct contact constraints of rim entry and in-hole insertion, separate PPO policies are trained for the two stages. Similar to our method, PPO first summarizes bilateral tactile feedback and robot-motion information into a low-dimensional state and then performs closed-loop action selection. The rim-entry PPO addresses position and orientation adjustment under peg--rim contact and uses a full six-degree-of-freedom Cartesian action space including both translation and rotation. The in-hole PPO addresses constrained advancement after the peg tip has entered the hole; its compact state consists of bilateral contact-force summaries, action-transfer effectiveness, accumulated advancement, and the previous Cartesian action, and the policy outputs three-dimensional Cartesian translations. Both PPO policies are trained independently in MuJoCo. For the end-to-end execution, the two policies are connected by a simple force-based transition detector. During rim contact, axial preload is maintained against the hole rim; when the peg enters the hole, the abrupt change in contact constraint produces a sharp decrease in the measured axial contact force. A fixed threshold on this force drop is used as the operational signal for switching from the rim-entry policy to the in-hole policy. Because non-convex peg--hole geometries are approximated by multiple convex collision bodies in simulation, the nominal CAD size difference does not directly represent the effective clearance under collision constraints. The critical jamming boundary of the simulated collision model is therefore first identified, and the effective training clearance is set to \(0.4\,\mathrm{mm}\) above this boundary.

The overall-performance results use online task completion within the applicable step budget without object release. The subsequent cue-validation and ablation analyses use additional offline reference, stability, and motion-transfer measures that do not affect controller execution.

\subsection{Overall Jamming-Preventive Performance}

Table~\ref{tab:real_success_rates} reports the success rates of rim entry, in-hole insertion, and end-to-end assembly across the \(18\) conditions, with \(20\) independent trials per method, condition, and protocol.

\begin{table}[t]
\centering
\scriptsize
\setlength{\tabcolsep}{1.5pt}
\renewcommand{\arraystretch}{1.18}
\vspace{3mm}
\caption{\textbf{REAL-ROBOT SUCCESS RATES ACROSS SHAPES, SIZES, CLEARANCES, AND PROTOCOLS.}}
\label{tab:real_success_rates}

\resizebox{1.0\columnwidth}{!}{%
\begin{tabular}{lcccccccccccc}
\toprule

\noalign{\global\setlength{\tabcolsep}{0pt}}
\multirow{2}{*}{\makecell[c]{Setting}} 
& \multicolumn{2}{c}{\hc{condblue}{20mm Cyl.}} 
& \multicolumn{2}{c}{\hc{condblue}{20mm Square}} 
& \multicolumn{2}{c}{\hc{condblue}{20mm Tri.}} 
& \multicolumn{2}{c}{\hc{condblue}{16mm Cyl.}} 
& \multicolumn{2}{c}{\hc{condblue}{16mm Square}} 
& \multicolumn{2}{c}{\hc{condblue}{16mm Tri.}} \\
\cmidrule(lr){2-13}

& \hc{oursblue}{ours} & \hc{ppopink}{PPO}
& \hc{oursblue}{ours} & \hc{ppopink}{PPO}
& \hc{oursblue}{ours} & \hc{ppopink}{PPO}
& \hc{oursblue}{ours} & \hc{ppopink}{PPO}
& \hc{oursblue}{ours} & \hc{ppopink}{PPO}
& \hc{oursblue}{ours} & \hc{ppopink}{PPO} \\
\midrule

\noalign{\global\setlength{\tabcolsep}{1.5pt}}

\multicolumn{13}{l}{\textit{0.5\,mm clearance}} \\
\midrule
Alignment 
& 1.00 & 0.40 & 0.90 & 0.15 & 0.95 & 0.20 & 0.95 & 0.45 & 0.90 & 0.15 & 0.95 & 0.15 \\
Insertion 
& 1.00 & 0.55 & 1.00 & 0.60 & 1.00 & 0.65 & 1.00 & 0.50 & 1.00 & 0.55 & 1.00 & 0.40 \\
\makecell[l]{End2end}
& 1.00 & 0.20 & 0.90 & 0.10 & 0.95 & 0.00 & 0.90 & 0.20 & 0.85 & 0.05 & 0.85 & 0.05  \\
\midrule

\multicolumn{13}{l}{\textit{0.4\,mm clearance}} \\
\midrule
Alignment 
& 1.00 & 0.30 & 0.90 & 0.10 & 1.00 & 0.10 & 1.00 & 0.25 & 0.90 & 0.00 & 0.90 & 0.00 \\
Insertion 
& 1.00 & 0.55 & 1.00 & 0.60 & 1.00 & 0.60 & 1.00 & 0.40 & 1.00 & 0.55 & 1.00 & 0.60 \\
\makecell[l]{End2end}
& 1.00 & 0.20 & 0.90 & 0.00 & 0.90 & 0.00 & 0.80 & 0.00 & 0.90 & 0.10 & 0.90 & 0.00 \\
\midrule

\multicolumn{13}{l}{\textit{0.3\,mm clearance}} \\
\midrule
Alignment 
& 0.90 & 0.15 & 0.80 & 0.05 & 0.85 & 0.00 & 0.95 & 0.10 & 0.85 & 0.00 & 0.80 & 0.00 \\
Insertion 
& 1.00 & 0.35 & 0.90 & 0.40 & 0.95 & 0.30 & 1.00 & 0.30 & 0.95 & 0.35 & 0.90 & 0.25 \\
\makecell[l]{End2end}
& 0.90 & 0.05 & 0.75 & 0.05 & 0.85 & 0.00 & 0.85 & 0.05 & 0.85 & 0.00 & 0.90 & 0.00 \\

\bottomrule
\end{tabular}
}%
\vspace{-3.5mm}
\end{table}

Across the \(18\) conditions, the proposed method achieves higher observed success rates than the PPO baseline in all three protocols, with the advantage retained as clearance decreases. As clearance tightens, residual mismatch more readily produces constrained lateral contact and frictional resistance; the proposed tactile regulation helps preserve effective advancement under these stronger constraints.

Fig.~\ref{fig:exp_results} shows representative execution processes of the proposed method and the PPO baseline. Representative PPO failure cases include failure to establish a favorable entry contact and inappropriate lateral adjustment during constrained insertion, which may lead to poor advancement or object release.

\subsection{Validation of the Force--Position Coupled Tactile Cue}

This subsection evaluates the physical consistency of the two tactile-cue components using offline directional and force-related references.

For the directional component, we collect timestamp-aligned tactile and visual data under \(20\,\mathrm{mm}\) workpieces and a \(0.4\,\mathrm{mm}\) clearance for cylindrical, square-prismatic, and triangular-prismatic pegs. The principal directions \(\theta_t^{(L)}\) and \(\theta_t^{(R)}\) are extracted from the bilateral tactile fields and fused into \(e_{\theta,t}\) after applying the common bilateral coordinate convention and resolving the \(\pi\)-periodic ambiguity. The external camera and ArUco markers provide an offline roll-trend reference and do not participate in closed-loop control. Temporal consistency is evaluated using trial-level Spearman correlation and fixed-polarity \(1\)-s change-direction agreement. Across \(18\) valid timestamp-aligned trials, the Spearman correlation between \(e_{\theta,t}\) and the visual trend was positive in all trials, with a median of \(0.780\); the median \(1\)-s change-direction agreement was \(0.696\). These results show that the directional cue consistently tracks the temporal evolution of contact direction.

For the moment-sensitive component, the bilateral centroid difference relative to the initial grasp preload is used as the offline validation quantity,
\[
c_{\mathrm{mom},t}^{\mathrm{val}}
=
e_{x,t}-b_x,
\]
where \(b_x\) is a trial-specific constant offset measured under the initial preload. Because \(b_x\) is constant within each trial, this centering does not change the trial-level Spearman rank correlation. Based on the bilateral sensor-coordinate transformations, a reduced moment-related force factor that does not contain the contact centroids is further constructed,
\[
q_{y,t}
=
F_{x,L,t}^{\mathrm{sens}}
-
F_{x,R,t}^{\mathrm{sens}}.
\]
The sign and monotonic consistency between \(c_{\mathrm{mom},t}^{\mathrm{val}}\) and \(q_{y,t}\) are then examined.

After removing one byte-identical duplicate from the historical analysis set, \(53\) unique trials were retained. The median trial-level Spearman correlation was \(-0.560\), and \(50/53\) trials exhibited the expected inverse relationship. This result shows that the bilateral centroid asymmetry remains sensitive to the corresponding contact-moment tendency without requiring numerical moment reconstruction. A post-hoc sensitivity analysis further examined trials with more balanced bilateral loading and stable preload, where the same inverse relationship remained descriptively stronger; the filtering criteria, subset composition, and detailed statistics are reported in the supplementary material.

Together, these results show that the directional cue captures the evolution of contact-geometric direction, while the bilateral centroid asymmetry provides a moment-sensitive contact response for the subsequent cue-guided correction.

\begin{figure}[t]
    \vspace{3mm}
    \centering
    \includegraphics[width=\columnwidth]{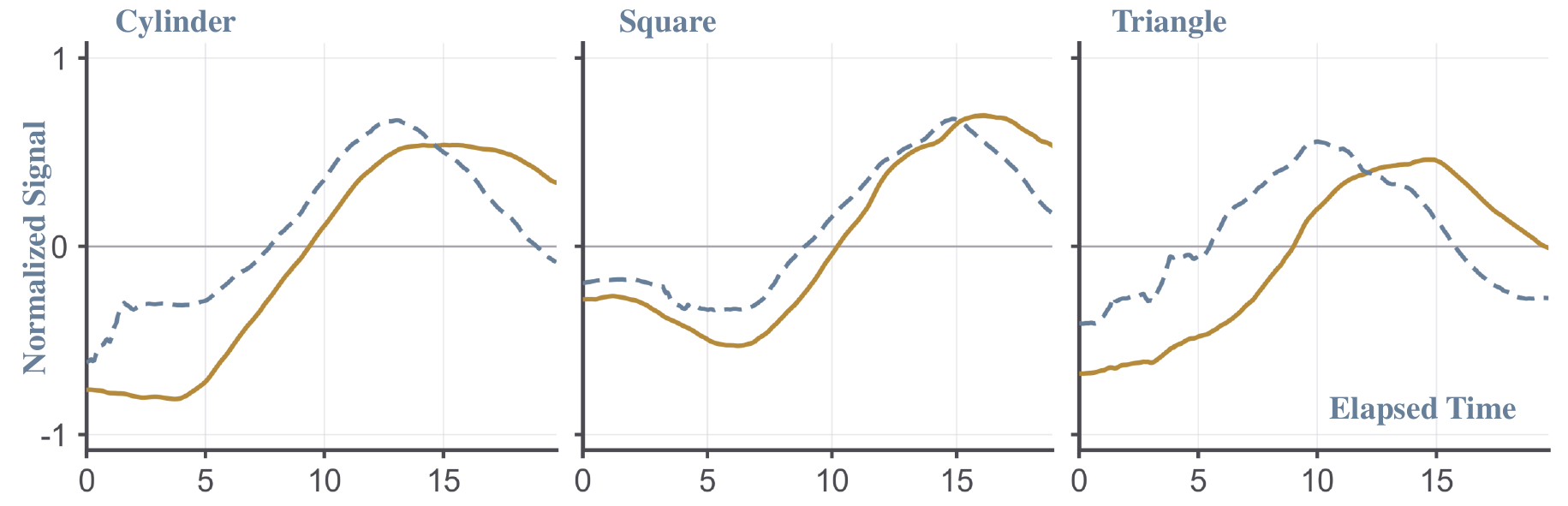}
    \caption{\textbf{Representative temporal consistency between tactile cues and offline references.}
    The directional cue is compared with the ArUco roll trend (top), and the bias-corrected centroid cue with the negated reduced moment-related force factor (bottom).}
    \label{fig:exp2_cue_validation}
    \vspace{-5mm}
\end{figure}

\subsection{Ablation on Dynamic Correction Regulation}

\begin{figure}[t]
    \vspace{3mm}
    \centering
    \includegraphics[width=\columnwidth]{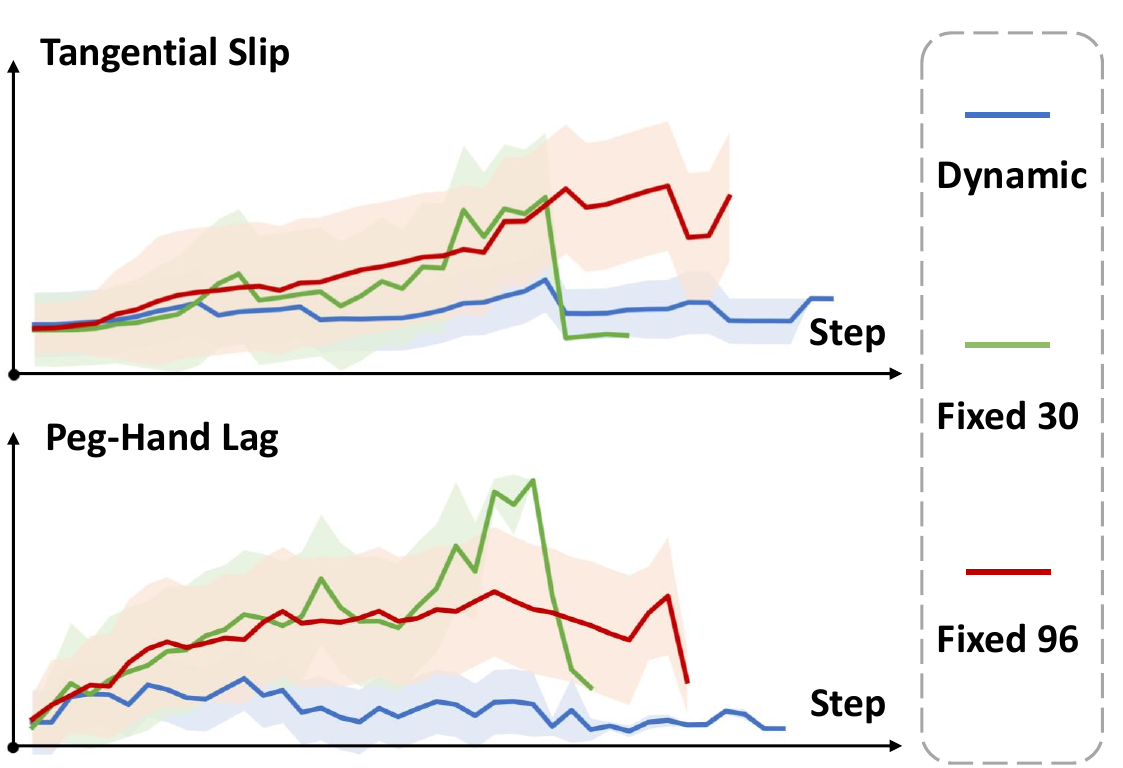}
    \caption{\textbf{Dynamic-regulation ablation.}
    Transverse contact-centroid drift and peg-to-hand axial lag are compared for Dynamic, Fixed-30, and Fixed-96.}
    \label{fig:ablation_results}
    \vspace{-5mm}
\end{figure}

This subsection examines whether regulating lateral correction according to tactile-cue evolution improves insertion stability and motion transfer. To isolate this mechanism, the tactile cue, the mapping from cue sign to correction direction, the low-level motion executor, and all other control parameters are kept unchanged. The complete Dynamic strategy is compared with two fixed-regulation variants, Fixed-30 and Fixed-96. Dynamic updates the target residual ratio according to the action-induced evolution of the tactile cue, whereas Fixed-30 and Fixed-96 keep the corresponding residual ratio fixed at \(\rho=0.30\) and \(\rho=0.96\), respectively.

The ablation uses \(20\,\mathrm{mm}\) workpieces, a \(0.4\,\mathrm{mm}\) clearance, and three peg shapes. For each strategy and each shape, we conduct \(6\) real-robot trials, giving \(18\) trials per strategy. The active controller is allowed to execute until task completion, the \(200\)-step limit, or manual termination after object release. Contact-force and centroid-drift thresholds are applied only offline to classify whether a completed trial remained stable.

Two offline instability events are defined. First, the bilateral force imbalance at macro step \(k\) is
\[
\Delta F_z(k)
=
\left|
F_{z,L}(k)-F_{z,R}(k)
\right|.
\]
A force-imbalance event is identified when
\[
\Delta F_z(k)>8\,\mathrm{N}
\]
for at least three consecutive logged macro steps. The threshold is at the nominal grasp-force scale and indicates a severe loss of balanced bilateral support.

Second, the transverse change of the bilateral contact-centroid separation is measured as
\[
D_y(k)
=
\left|
\left|c_{Ly}(k)-c_{Ry}(k)\right|
-
\left|c_{Ly}(0)-c_{Ry}(0)\right|
\right|.
\]
A transverse-drift event is identified when
\[
D_y(k)>6
\]
tactile-grid units at any logged macro step. This stability measure uses the sensor-frame \(y\) centroid component, orthogonal to the sensor-frame \(x\) component used by the online centroid cue \(e_{x,t}\). It therefore measures grasp-transverse contact redistribution. Object release is manually terminated for safety and is always counted as failure.

A trial is classified as a stable completion only if the insertion task is completed and neither of the above instability events occurs. These stability criteria are evaluated offline and do not alter the controller execution. In addition, external vision measures the peg-to-hand displacement along the insertion direction. The terminal axial lag is reported for every trial, and the mean terminal lag over all \(18\) trials of each strategy is used as a continuous measure of axial motion-transfer loss.

\begin{table}[t]
\centering
\scriptsize
\setlength{\tabcolsep}{3pt}
\renewcommand{\arraystretch}{1.18}
\vspace{3mm}
\caption{\textbf{STABLE COMPLETION RATE AND PEG-TO-HAND AXIAL LAG FOR DYNAMIC AND FIXED-REGULATION VARIANTS.}}
\label{tab:ablation_summary}

\resizebox{1.0\columnwidth}{!}{%
\begin{tabular}{lccc}
\toprule
Strategy config & Dynamic & Fixed-30 & Fixed-96 \\
\midrule
Stable completion rate & 0.83 & 0.56 & 0.67 \\
\midrule
Peg-to-hand axial lag(mm) & 1.25 & 3.12 & 4.40 \\
\bottomrule
\end{tabular}
}%
\vspace{-5mm}
\end{table}

As shown in Table~\ref{tab:ablation_summary}, Dynamic achieves a stable completion rate of \(0.83\), compared with \(0.56\) for Fixed-30 and \(0.67\) for Fixed-96. The mean terminal peg-to-hand axial lag of Dynamic is \(1.25\,\mathrm{mm}\), compared with \(3.12\,\mathrm{mm}\) for Fixed-30 and \(4.40\,\mathrm{mm}\) for Fixed-96. Thus, Dynamic exhibits both a higher observed stable-completion rate and a smaller terminal loss between commanded hand advancement and actual peg advancement.

Fig.~\ref{fig:ablation_results} uses three rows to show Dynamic, Fixed-30, and Fixed-96, and two columns to show the temporal evolution of transverse contact-centroid drift and peg-to-hand axial lag. The two fixed configurations cannot adapt correction aggressiveness to changing contact constraints and can therefore become either insufficient or excessive at different stages. In contrast, Dynamic preserves correction capability while the tactile cue continues to improve and attenuates subsequent correction when the cue stagnates or rebounds. The resulting reduction in transverse instability and terminal axial lag supports the role of action-induced tactile evolution in maintaining effective motion transfer before jamming forms.

\section{CONCLUSION}

This work presented a tactile regulation strategy for jamming-preventive precision peg-in-hole assembly. A force--position coupled tactile cue captures directional contact deviation and moment tendency from bilateral tactile fields, while action-induced cue evolution regulates lateral correction during maintained axial advancement. Stage-specific rim-entry and in-hole control instantiate this principle under changing contact constraints. Real-robot experiments across multiple shapes, sizes, and sub-millimeter clearances show improved task success over PPO baselines, while cue validation and ablation results support the physical consistency of the tactile representation and the effectiveness of dynamic correction regulation. The results show that evaluating the contact response induced by corrective actions provides an effective basis for maintaining motion transmission and suppressing jamming during precision insertion.










\bibliographystyle{IEEEtran}
\bibliography{root}

\end{document}